\documentclass[letterpaper, 10 pt, conference]{ieeeconf}  

\IEEEoverridecommandlockouts                              

\usepackage{graphics} 
\usepackage{epsfig} 
\usepackage{mathptmx} 
\usepackage{times} 
\usepackage{amsmath} 
\usepackage{amssymb}  
\usepackage{booktabs} 
\usepackage{tikz}
\usetikzlibrary{positioning,shapes.geometric,arrows,fit,shapes.symbols,svg.path,tikzmark,calc}
\usepackage{booktabs}
\usepackage{textcomp}
\usepackage{listings}

\pgfdeclarelayer{background}
\pgfdeclarelayer{foreground}
\pgfsetlayers{background,main,foreground}
\usepackage{siunitx}
\usepackage[short]{optidef}
\let\labelindent\relax
\usepackage[inline]{enumitem}
\usepackage{multirow}
\usepackage{multicol}
\usepackage{adjustbox}
\usepackage{float}

\usepackage{xcolor}
\usepackage{xspace}
\usepackage{float}
\usepackage{url}
\usepackage{hyperref}
\usepackage{makecell}
\newcommand{\algabbr}{Omni-Scan\xspace}

\hypersetup{
    colorlinks=true,       
    linkcolor=magenta,     
    urlcolor=magenta       
}

\title{\LARGE \bf
\algabbr: Creating Visually-Accurate Digital Twin Object Models Using a Bimanual Robot with Handover and Gaussian Splat Merging 
}

\author{Tianshuang Qiu$^{*1}$, Zehan Ma$^{*1}$, Karim El-Refai$^{*1}$, Hiya Shah$^{1}$, Chung Min Kim$^{1}$, Justin Kerr$^{1}$, Ken Goldberg$^{1}$ 
\thanks{$^{*}$ Equal contribution}%
\thanks{$^{1}$University of California, Berkeley
}
}

\usepackage{caption}

\renewcommand{\paragraph}[1]{\noindent\textbf{#1}}
\usepackage{float}

\begin{document}

\maketitle
\thispagestyle{empty}
\pagestyle{empty}

\begin{abstract}
3D Gaussian Splats (3DGSs) are 3D object models derived from multi-view images. Such ``digital twins'' are useful for simulations, virtual reality, E-commerce, robot policy fine-tuning, and part inspection. 3D object scanning usually requires multi-camera arrays, precise laser scanners, or robot wrist-mounted cameras, which have restricted workspaces. We propose \algabbr, a pipeline for producing high-quality 3D Gaussian Splat models using a bi-manual robot that grasps an object with one gripper and rotates the object with respect to one stationary camera. The object is then re-grasped by a second gripper to expose surfaces that were occluded by the first gripper. We present the \algabbr{} robot pipeline using DepthAnything, Segment Anything, as well as RAFT optical flow models to identify and isolate objects held by a robot gripper while removing the gripper and the background. We then modify the 3DGS training pipeline to support concatenated datasets with gripper occlusion, producing an omni-directional (360$^\circ$) model of the object. We apply \algabbr{} to part defect inspection, finding that it can identify visual or geometric defects in 12 different industrial and household objects with an average accuracy of 83.3\% . More details and interactive videos of \algabbr 3DGS models can be found at \url{https://berkeleyautomation.github.io/omni-scan/}.
\end{abstract}
\begin{figure}[t!]
    \centering
    \includegraphics[width=0.9\linewidth]{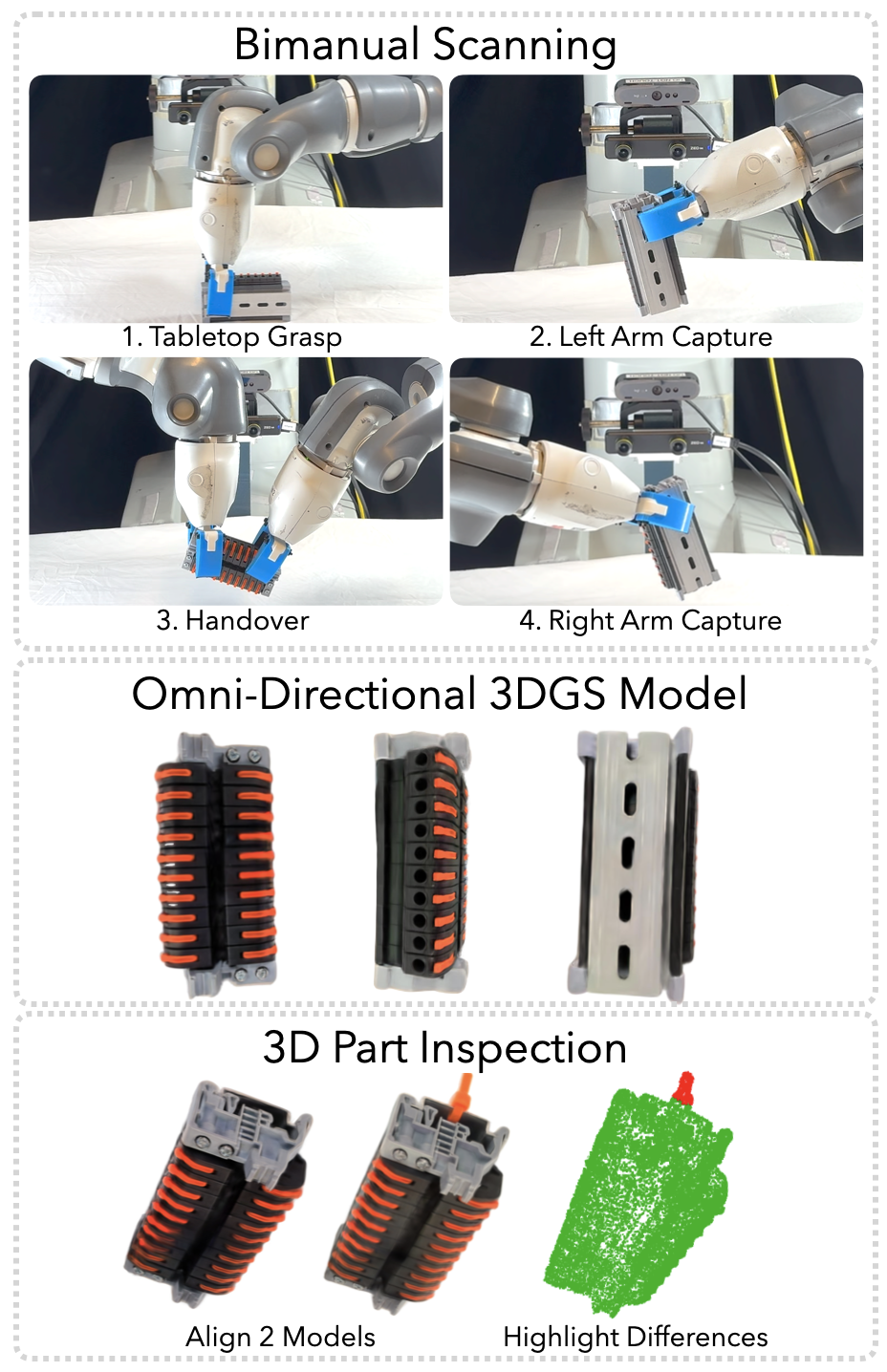}
    \caption{The robot grasps an object (wire connector) in any position and orientation for inspection. \algabbr then transfers the object between grippers to create a complete scan. The resulting full surface 3DGS model can be compared with a reference model for object inspection.}
    \vspace{-1em}
    \label{fig:splash}
\end{figure}

\begin{figure*}[t]
    \centering
    \includegraphics[width=0.95\linewidth]{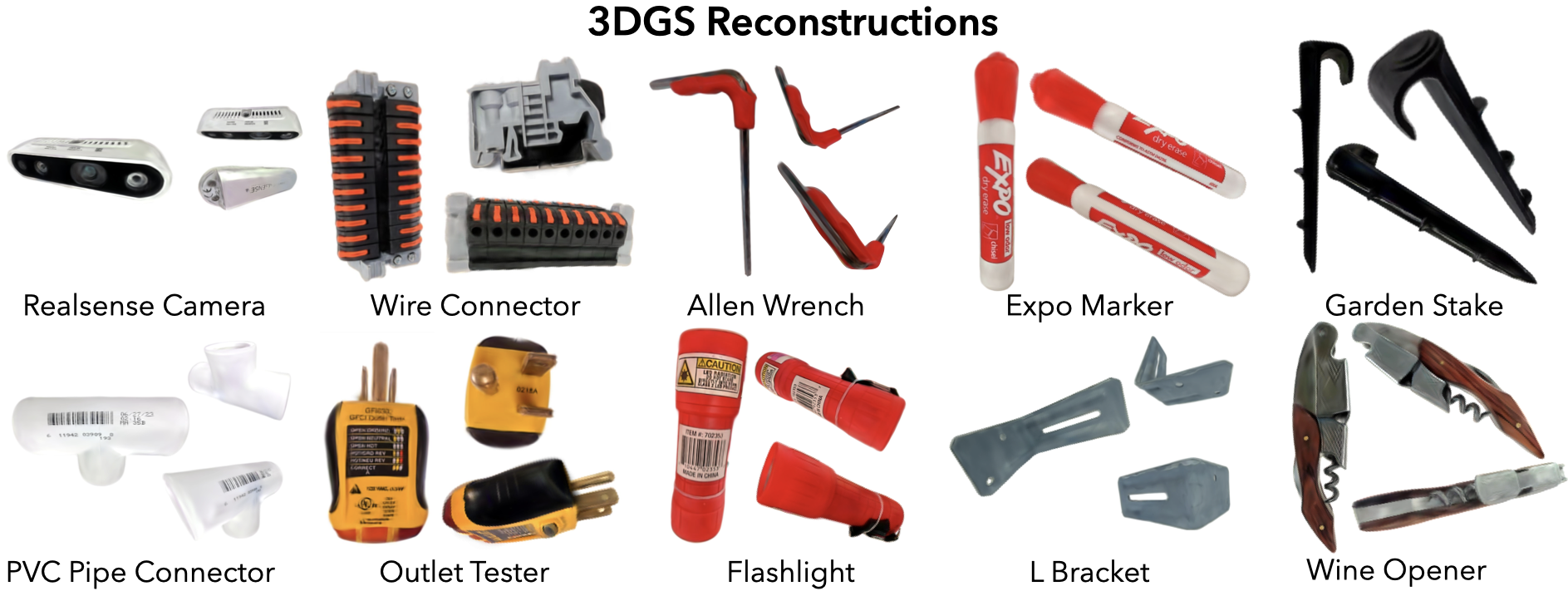}
    \caption{\footnotesize \textbf{Reconstructed 3D Gaussian Splats of the 3DGS-Merged Model} We show \textbf{rendered views} from reconstructed splat models of objects collected by \algabbr{}. Each object is fully reconstructed without occlusion, even though the data was collected while grasped. In addition, the models capture fine geometric and visual details such as text or notches. See \href{https://berkeleyautomation.github.io/omni-scan/}{our website} for interactive videos of full 3D surfaces.}
    \label{fig:reconstructions}
    \vspace{-2em}
\end{figure*}
\section{Introduction}
Most 3D object scanning methods rely on multiple fixed cameras. 
Recent advances in 3D reconstruction, such as Neural Radiance Fields (NeRF) \cite{mildenhall2020nerf} and 3D Gaussian Splatting \cite{kerbl2023gaussian}, have enabled high-quality novel view synthesis and 3D reconstruction from multiple 2D images. However, in robotic contexts, prior work has used moving wrist-mounted cameras, which significantly limits coverage of the object due to kinematic arm constraints and inability to scan sections of the object surface near the support surface.

In this paper, we present \algabbr{}, a fully autonomous system for 3D object reconstruction through in-hand scanning with a bi-manual robot, using only one stationary RGB camera and a stereo depth sensor.
The robot grasps and rotates objects in front of the camera, capturing comprehensive multi-angle views.
To expose surfaces occluded by the gripper, the system incorporates a bi-manual handover process that repositions the object for full 360° coverage and generates an omni-directional 3D Gaussian Splatting (3DGS) model.
In-gripper scanning presents unique challenges, such as occlusions from the end effector, merging scans from multiple grasps, and the inversion of the typical neural reconstruction assumption (i.e., a moving camera capturing a static scene). To overcome this, we design a masking pipeline that segments the gripper, object, and background using optical flow, DepthAnything V2~\cite{NEURIPS2024_26cfdcd8}, Segment Anything \cite{kirillov2023segment}, and Segment Anything 2~\cite{ravi2025sam} for each view. We then alter the traditional 3D Gaussian Splatting training pipeline to accommodate this new scanning paradigm. We apply \algabbr to industrial part inspection, where it identifies both visual and geometric defects in household and industrial objects from a reference object.

This paper makes the following contributions:
\begin{enumerate}
    \item \algabbr: A pipeline for bi-manual robot object scanning that includes grasping, multi-view scanning, handover for complete coverage, and 3D model generation.
    \item A robust masking and processing approach that accurately distinguishes the object, gripper, and background in captured images.
    \item A pose optimization and model merging technique that aligns and combines multiple 3DGSs into a cohesive 3D Gaussian Splat.
    \item A novel method for aligning 3DGS models for defect inspection.
    \item Experimental results evaluating the effectiveness of \algabbr for visualization and part inspection applications, achieving an accuracy of 83\% for defect detection.
\end{enumerate}

\begin{figure*}[t!]
    \centering
    \includegraphics[width=0.95\linewidth]{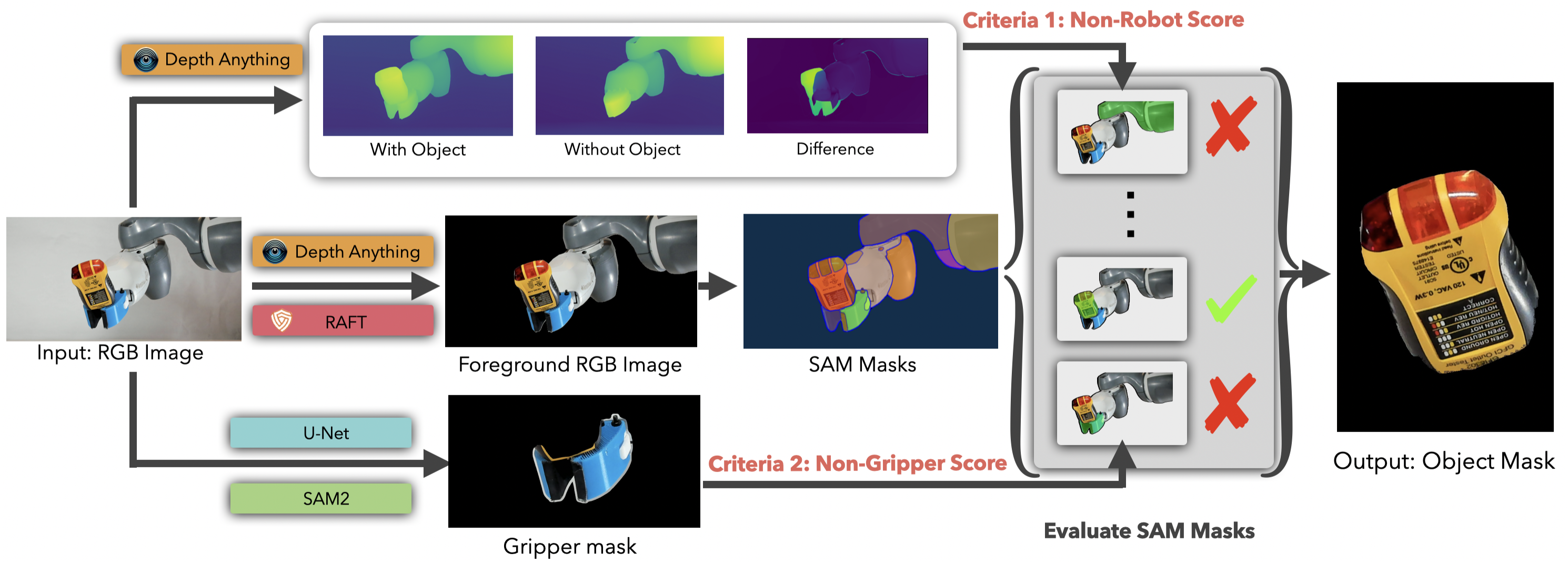}
    \caption{\footnotesize \textbf{\algabbr Masking Pipeline} (1) starts with an RGB image of the robot gripper holding an object, (2) extracts the foreground to isolate potential objects, (3) uses SAM to generate candidate object masks, (4) evaluates masks using two criteria: Non-Robot Score (comparing depth with/without object) and Non-Gripper Score (using U-Net and SAM2-generated gripper masks), and (5) outputs a clean object mask containing only the target object, rejecting gripper and robot parts. Note: Overlayed RGB masks are used for illustration; the actual pipeline outputs binary masks.
}
    \label{fig:masking}
    \vspace{-2em}
\end{figure*}
\section{Related Work}

\subsection{3D Reconstruction with Radiance Fields}
Neural Radiance Fields ~\cite{mildenhall2020nerf} are an attractive representation for high quality scene reconstruction from posed RGB images, with a flurry of recent work enhancing quality~\cite{adamkiewicz2022vision,barron2021mip,barron2022mip,ma2022deblur}, large-scale scenes~\cite{tancik2023nerfstudio,wang2023f2,barron2023zip}, optimization speed~\cite{muller2022instant,Chen2022ECCV,fridovich2023k,fridovich2022plenoxels}, dynamic scenes~\cite{park2021hypernerf,li2023dynibar,pumarola2020d}, and more. Because of its high-quality reconstruction and differentiable properties, NeRF has been explored in robotics for navigation and mapping~\cite{adamkiewicz2022vision,Zhu_2022_CVPR,Sucar_2021_ICCV,rosinol2023nerf}, manipulation~\cite{li20223d,22-driess-NeRF-RL,kerr2022evonerf,IchnowskiAvigal2021DexNeRF,Rashid2023LanguageER,kerr2024rsrd,shen2023F3RM}, and for synthetic data generation~\cite{byravan2023nerf2real}. 3D Gaussian Splatting ~\cite{kerbl2023gaussian} made a major breakthrough in speed and quality of radiance fields, and the field has quickly adopted it for similar applications. In this work we use 3DGS to reconstruct high-quality object models, and in contrast to prior work reconstruct entire objects in high detail with a static camera, via a method of merging multiple scans and accurately masking the object of interest.

\subsection{3D Object Scan Data}
Conventionally, datasets of 3D objects are constructed with expensive equipment like multiview camera arrays or high precision depth sensors, such as in the Google Scanned Objects~\cite{downs2022google} or DTU~\cite{aanaes2016large} datasets. Other large datasets like Objaverse~\cite{deitke2023objaverse} exist, but are comprised of synthetic objects. In this work, we leverage recent work on multi-view reconstruction from RGB images to alleviate the need for expensive sensors and autonomously digitize real objects with a robot.

Several works explore reconstructing objects in human hands, including Color-NeuS~\cite{zhong2024color}, which reconstructs object SDFs by separating view-dependent effects with a relighting network. BundleSDF~\cite{wen2023bundlesdf} achieves near real-time tracking and reconstruction from monocular RGB-D video through pose graph optimization. 


\subsection{Automated Part Inspection}
Automated part inspection using robotics has advanced significantly with vision systems, machine learning, and sensor integration. Prior work has studied photogrammetry-based 3D reconstruction for inspection where the robot moves a camera around the object on a tabletop~\cite{khan2021vision}.
Davtalab et al. (2022) developed a deep learning approach for real-time defect detection in additive manufacturing, improving quality control~\cite{davtalab2022automated}

\section{Problem Statement}
The goal is to create a visually accurate, omni-directional 3D model of a given object, and use it for defect inspection. We assume objects are rigid and cannot fit inside a 3cm diameter sphere but can fit inside a 10cm one, as well as the availability of a bi-manual robot with parallel jaw grippers, one fixed high-resolution monocular camera, and one stereo camera. During reconstruction, a target object is placed within the robot's reachable workspace on a tabletop. We assume the robot is able to grasp and lift the object (i.e it is not too heavy). During defect inspection, a robot is provided with 3DGS models of two reference objects and one new 3DGS model to evaluate. The system analyzes these 3 models to determine if the new model contains a defect and if so, where. Defects are geometric defects, meaning a structural deformation or flaw greater than 4.5mm in size, or visual defects, such as a scratch or a blemish greater than 2mm in size.

\begin{figure*}
    \centering
    \includegraphics[width=0.95\linewidth]{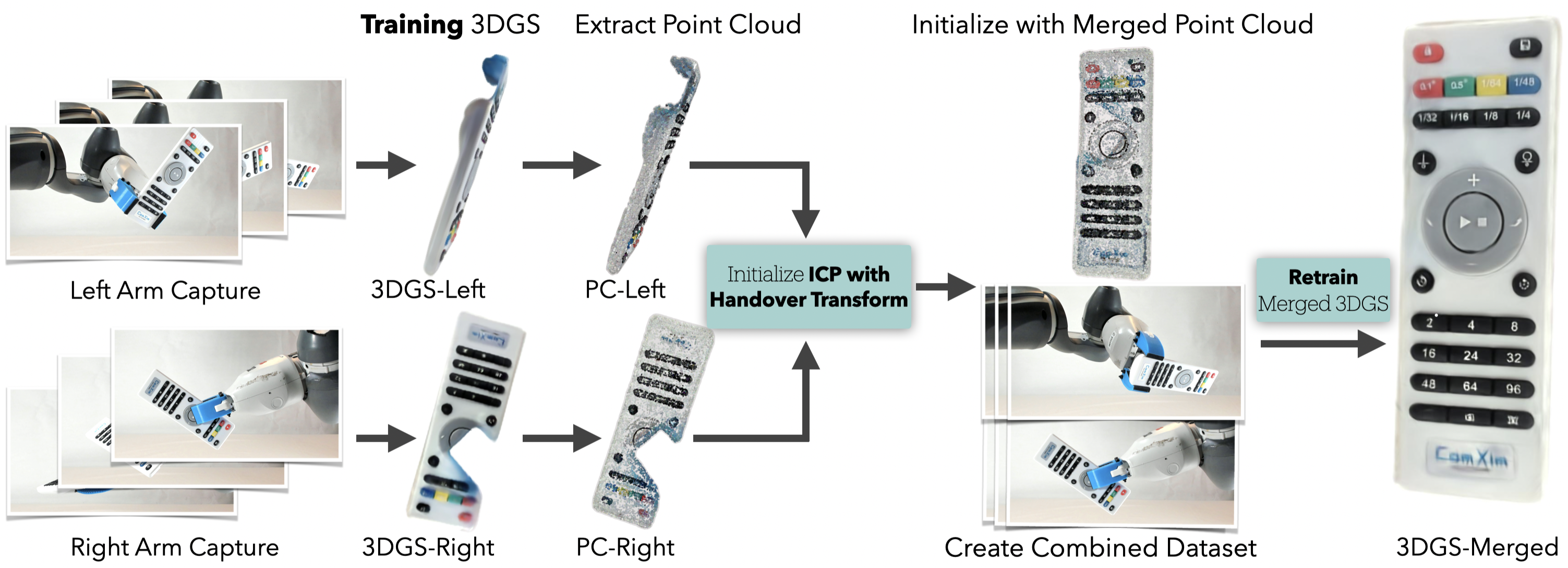}
    \caption{\footnotesize \textbf{Overview of Training Pipeline} We first train separate 3DGS models for left and right arm captures and extract their Gaussian centers as point clouds. Using the estimated handover transform \(T_{lr}\), we initialize Iterative Closest Point (ICP) algorithm, which iteratively refines the alignment between two point clouds by minimizing the distance between corresponding points, for alignment. The refined transformation from ICP is then used to merge the datasets, enabling training of a unified 3DGS model on the combined dataset.}
    \label{fig:merged}
    \vspace{-2em}
\end{figure*}
\section{\algabbr{}}

\algabbr begins by grasping the object from the tabletop and rotating it mid-air in front of a fixed camera to capture multi-view images. We then perform a handover, passing the object from one gripper to another to scan it again from a new pose. After collecting the images, we process them with a combination of robot kinematics, Depth Anything, optical flow, and SAM to generate training poses and masks for 3DGS reconstruction. We then train 2 individual Gaussian Splat models (left and right) and merge them into a single, high-quality 3DGS model. We use the resulting model for part inspection by detecting defects compared to other examples of the same object.

\subsection{Scanning Procedure}
\subsubsection{Tabletop Grasping}
\label{methods:grasping}
We use an ABB YuMi bi-manual robot with soft 3D-printed grippers~\cite{elgeneidy2019gripper} for compliant caging grasps. Objects are placed randomly within the workspace and captured by a ZED Mini RGB-D camera and a Logitech BRIO RGB camera mounted at the robot base.

We generate a depth image of the object from the ZED Mini's stereo image pairs using RAFT-Stereo~\cite{shankar2022learned} and one RGB image to generate object masks with SAM~\cite{kirillov2023segment}, filtering the masks by the known location of the table to isolate the object. The depth image is deprojected to create point clouds of both the scene and the isolated object, using DBSCAN~\cite{ester1996density} to remove noise. Contact-GraspNet~\cite{sundermeyer2021contact} then generates candidate grasps on only the object point cloud, and the highest-scoring grasp is planned and executed with the left side gripper using the Jacobi motion planning software \cite{jacobi2024motion}. If the grasp is kinematically infeasible or would lead to a collision, the next highest scored grasp is chosen.

\subsubsection{Scan Trajectory}
\label{methods:scanning}
After the object is grasped and lifted, the robot performs scanning by rotating the wrist of the gripper $360^{\circ}$ in 20 evenly spaced longitudinal positions about its local z-axis. We evenly sample 5 latitudes from the z-axis between -10 and 70 degrees (equaling 100 images). Beyond these limits, occlusions from the gripper prevent the camera from clearly viewing the object. At each latitude, we collect the pose of the arm that is holding the object $T$ and capture the corresponding 4K image $I$ from the Logitech BRIO. The scanning process for one arm takes 6 minutes for 100 images.

\subsubsection{Bi-Manual Handover}
Since part of the object is occluded by the gripper during the initial left scan, the robot performs a bi-manual handover and scans again from a new pose to capture the previously hidden regions. Unlike single-arm regrasping, which places the object down and picks it up again, bi-manual handover provides a reliable relative transform between scans, which enables more accurate pointcloud registration when merging the right-arm scan into the left-arm coordinate frame, as shown in \ref{method: align_scans}. It is especially useful for elongated objects (e.g., Allen Wrench and Expo Marker in Fig. \ref{fig:reconstructions}), which are often unstable when placed on a planar worksurface on their endpoints.
To execute bi-manual handover, the robot moves the object to a predefined end-effector position easily reachable by the other arm.
Following a very similar approach as~\ref{methods:grasping}, \algabbr{} generates grasps on the object point cloud after segmenting the robot arm by deleting depth points overlapping with the URDF model. We then choose the highest scored grasp, accounting for kinematic constraints and collisions. To handover the object, the right gripper encloses the object, and then the left gripper is released. This right arm then repeats the same scanning process as detailed in \ref{methods:scanning}.

\subsection{Dataset Processing}
\subsubsection{Pose Processing}
\label{method:pose_processing}
We first compute the camera-to-object transform for the left and right scans (100 images per scan). From the calibrated camera, we can get the transform from camera to world $T_{c}$. Since we do not directly have the pose of an object center relative to the robot, we approximate it with the transform from the robot to the gripper. The reconstruction of the object is performed in the frame of the gripper.

For each image $i$, the pose from the camera to the object $T_{ic}$ can be computed by its corresponding $T_{i}^{-1}T_{c}$, where $T_{i}$ is the transform from the robot gripper (that is holding the object) to world, creating $\text{capture}_L$ and $\text{capture}_R$, consisting of image-transform pairs.

\subsubsection{Mask Processing}

The masking pipeline (Figure~\ref{fig:masking}) robustly segments the object by systematically filtering out background elements, robot gripper, and robot arm. The pipeline consists of the following key components:

\paragraph{Robot Gripper Segmentation}

We trained a U-Net model on 3,000 labeled images to obtain initial gripper masks. However, due to sensitivity to lighting and object colors, its predictions are occasionally inaccurate or include parts of the object. For consistent and accurate gripper masks, we hand-select a few frame indices--shared across all objects--where the gripper is clearly visible and unoccluded, and use their high-confidence U-Net gripper masks to prompt SAM2 for robust video-based gripper mask propagation 
\cite{ravi2025sam}.

\paragraph{Object Mask Generation}
To isolate the object from the robot and background, first we obtain a foreground mask by thresholding depth values to be small, and in addition keeping only pixels with non-zero optical flow between neighboring frames. This foreground mask is passed to SAM2 to generate a set of candidate masks. In addition, we process each frame through DepthAnything V2~\cite{NEURIPS2024_26cfdcd8}, which provides estimated per-pixel depth. We also perform a one-time calibration where we estimate mono-depth for images from the capture trajectory without an object grasped (empty gripper). The usage of these depth maps is described next.

For each candidate mask \( M \) from SAM, we then label it as part of the robot or object based on two scoring functions:

\textbf{Non-Robot Score}:
\begin{equation}
    S_{\text{NR}} = \frac{1}{|M|} \sum_{p \in M} |D_{curr}(p) - D_{empty}(p)|
\end{equation}
where \( D_\text{curr} \) is the current depth output, \( D_{empty} \) is the empty-gripper depth output, and \( p \) represents pixels in the candidate mask. Since the depth of the robot arm in current is typically similar to the empty-gripper depth, the score helps filter out regions corresponding to the arm. In contrast, the object and gripper configuration will differ significantly from the empty gripper depth (where the gripper is fully closed and no objects are in it), resulting in a higher \( S_\text{NR}\) score, indicating a higher likelihood of belonging to the object.

\textbf{Non-Gripper Score}
\begin{equation}
    S_{\text{NG}} = 1 - \frac{|M \cap G|}{|M|}
\end{equation}
where \( G \) is the gripper mask. A higher \( S_\text{NG}\) score indicates less overlap with the gripper, meaning it's more likely part of the object.

We keep candidate masks with \( S_{NR} \geq 140 \) and \( S_{NG} \geq 0.89 \) as the final object mask (threshold empirically determined).

\subsection{3DGS Training}
After obtaining the object masks, \algabbr{} seeks to create one omni-directional Gaussian Splat model of the entire object without occlusions. We do this in the following steps:
\begin{enumerate}
    \item Create $\text{capture}_L$ and $\text{capture}_R$ from the left and right arm scans
    \item Train Gaussian Splat models, $\text{3DGS}_L$ and $\text{3DGS}_R$, \textbf{individually} on $\text{capture}_L$ and $\text{capture}_R$
    \item Compute $\text{capture}_\text{merge}$ by computing equivalent transforms between $\text{capture}_L$ and $\text{capture}_R$
    \item Train $\text{3DGS}_\text{merged}$ on $\text{capture}_\text{merge}$ as the \textbf{merged} 3D model
\end{enumerate}

\subsubsection{Compute \texorpdfstring{$\text{capture}_L$}{capture	extunderscore L} and \texorpdfstring{$\text{capture}_R$}{capture	extunderscore R}}
Using the method outlined in section \ref{method:pose_processing}, we compute image-transform pairs for all individual scans. This transform is still in the respective grasp frame, so it is only suitable for training the individual models, $\text{3DGS}_L$ and $\text{3DGS}_R$.

\subsubsection{Training Individual Models}
\label{method: pc_extraction}
We first produce a 3DGS of each of the datasets individually for 16000 steps to get an estimate of the object's geometry. From these splat models, we retrieve colored point clouds $P_1, P_2$.

\subsubsection{Aligning the scans to create \texorpdfstring{$\text{capture}_\text{merge}$}{capture-merge}}

\label{method: align_scans}

To train a complete $\text{3DGS}$ model from both scans, we need to transform the right capture into the coordinate frame of the left capture, which we treat as the canonical frame. We aim to estimate the relative transform from the right gripper to the left at handover, and use it to align the right scan to the left scan’s frame.

Let the left and right gripper poses (gripper to world) at handover be $T_{lh}, T_{rh}$. Each image captured while the object is held by the left gripper is denoted $i^l$, and by the right gripper $i^r$. The corresponding pose of each image in its own camera frame is $T_{ic}^l$ or $T_{ic}^r$. We denote the fixed transform from the camera to the world as $T_{c}$. To express the handover transform in the camera frame, we convert both gripper poses into the camera frame via $T_c^{-1}T_{lh}$ and $T_c^{-1}T_{rh}$. The equivalent left-frame pose for any right image $i^r$ is then: 

\begin{align}
T_{ic} ^l &= (T_c^{-1}T_{lh})^{-1} T_c^{-1}T_{rh} T_{ic}^r \\
&= T_{lh}^{-1} T_c T_c^{-1}T_{rh} T_{ic}^r = T_{lh}^{-1}T_{rh} T_{ic}^r
\end{align}

This transformation $T_{lh}^{-1}T_{rh}$ gives an initial estimate of the rigid transform from right scan to left scan. However, due to imprecision in the handover poses, this estimate is often insufficiently accurate. To refine it, we apply the Iterative Closest Point (ICP) algorithm.

Specifically, we take colored point clouds $P_1, P_2$ from \ref{method: pc_extraction},  initialize ICP using the handover-based transform $T_{lh}^{-1}T_{rh}$, and run two variants: standard ICP and colored ICP. We then select the optimized transform $T_{lr}^*$ by comparing the fitness scores, choosing the one that achieves better alignment. The right image poses are updated as:  $T_{ic}^l =T_{lr}^* T_{ic}^r$, while left scan remains unchanged since it is the canonical frame.

\subsubsection{Training Omni-Directional Model on Merged Captures} Using the merged colored point clouds $P_1 + T^*_{lr}P_2$ as initialization for the 3DGS model, we train $\text{3DGS}_\text{merge}$ on $\text{capture}_\text{merge}$ for 50000 steps. 

\subsection{Supporting In-Gripper Datasets} 
 For 3DGS training, we extend Nerfstudio's Splatfacto model~\cite{nerfstudio,ye2024gsplatopensourcelibrarygaussian} to support multi-dataset training. Naively training a 3DGS on the raw image datasets is infeasible as 3DGS assumes a static scene, while our data seen from the perspective of the camera is inherently inconsistent except for the object. Thus, we must alter the losses to account for this. In addition, we must support training on datasets where the object is occluded by the gripper.

\paragraph{Object Opacity Loss} During the training process, Gaussian Splat models produce the \textbf{accumulation} metric as well as RGB renders. The accumulation metric measures how much each pixel is covered or influenced by overlapping Gaussians during the rendering process. Accumulation quantifies the total accumulated alpha (opacity) at each pixel due to the contribution of multiple Gaussians.
Lower accumulation values suggest sparse coverage, where fewer Gaussians contribute to the final pixel color. We introduce an L1 loss between the model's \textbf{accumulation} and the image's object mask, which attempts to match the rendered opacity to the calculated mask. Intuitively this penalizes any Gaussians outside of the object mask to ensure the resulting model is floater-free and has clean boundaries.

\paragraph{Gripper-Agnostic Losses} When combining the datasets, we formulate the loss such that the model is ambivalent towards the area that the gripper occupies. Specifically, any per-pixel loss value that intersects with a gripper mask is set to 0. Importantly, this includes the previously described opacity loss, which ensures the model is able to add Gaussians that are occluded by the gripper in one dataset by analyzing the object from the other dataset's perspective. 

\begin{figure}[h]
    \centering
    \includegraphics[width=\linewidth]{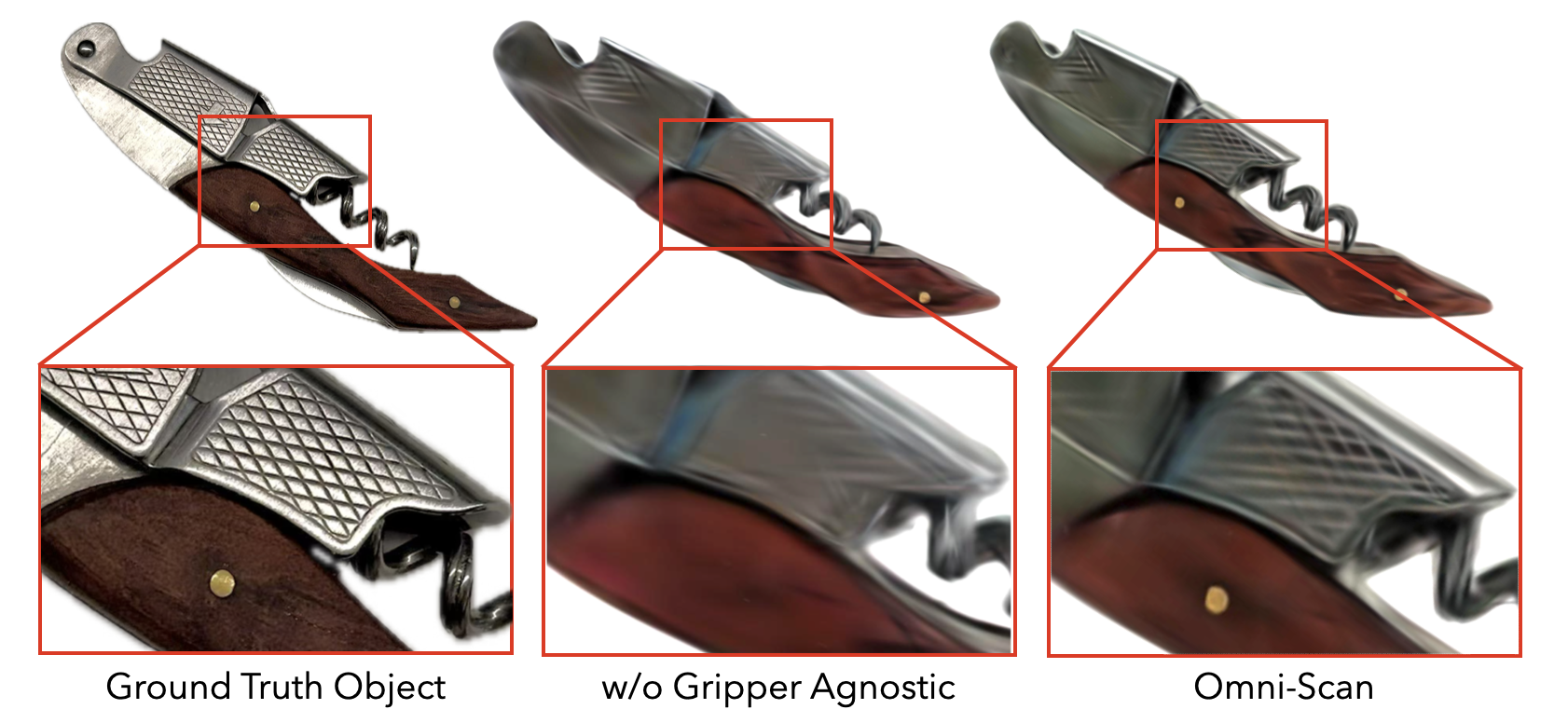}
    \caption{\textbf{Gripper Agnostic Loss Ablation} We perform an ablation on the Gripper Agnostic Loss, and we observe that reconstruction quality decreases without it. Specifically, the bronze stud and the cross-hatch pattern appear only when we have the Gripper Agnostic Loss.}
    \label{fig:gripper_excl}
    \vspace*{-0.2in}
\end{figure}
\begin{figure*}[ht]
    \centering
    \includegraphics[width=0.9\linewidth]{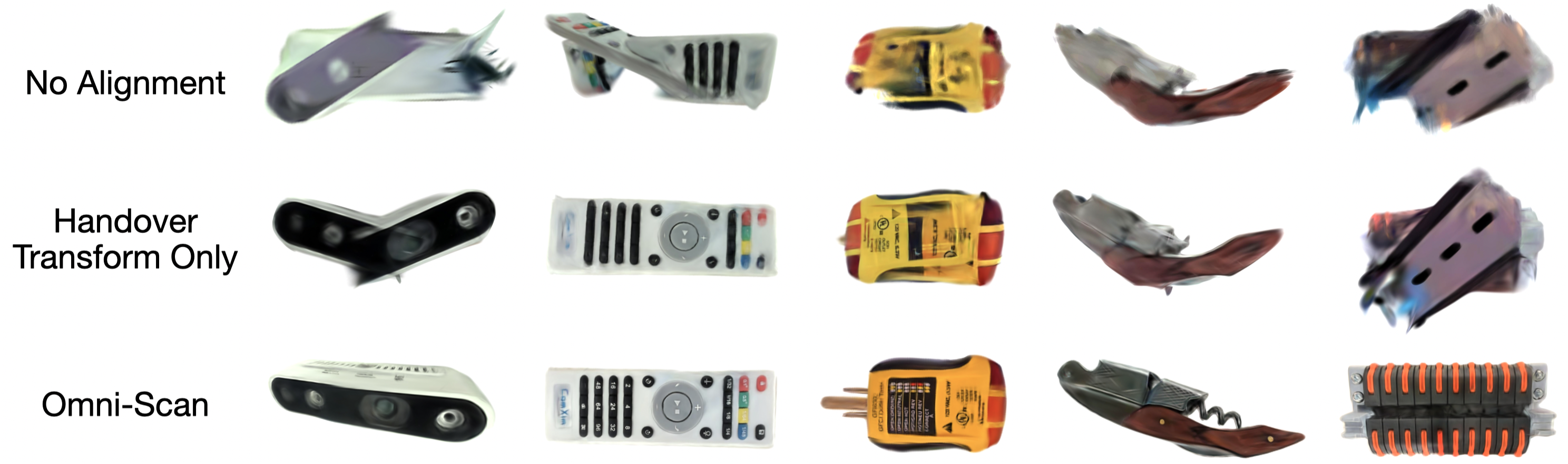}
    \caption{\textbf{Alignment Ablation} We perform an ablation on the alignment as illustrated in Fig. \ref{fig:merged} on 5 objects. We present renders for models with no alignment, with only handover transform, and Omni-Scan which uses the handover transform as an initialization for ICP alignment.}
    \label{fig:alignment-ablation}
    \vspace{-0.2in}
\end{figure*}
\setlength{\tabcolsep}{1.4pt}
\begin{table*}
\centering
\vspace{0.5em}
{\tiny
\begin{tabular}{lccccccccccccccc}
\toprule
  & \multicolumn{3}{c}{\textbf{Realsense Camera}} & \multicolumn{3}{c}{\textbf{Remote Control}} & \multicolumn{3}{c}{\textbf{Outlet Tester}} & \multicolumn{3}{c}{\textbf{Wine Opener}} & \multicolumn{3}{c}{\textbf{Wire Connector}} \\
  \cmidrule(lr){1-4} \cmidrule(lr){5-7} \cmidrule(lr){8-10} \cmidrule(lr){11-13} \cmidrule(lr){14-16}
  & \textbf{No Alignment} & \textbf{Handover Only} & \textbf{\algabbr{}} & \textbf{No Alignment} & \textbf{Handover Only} & \textbf{\algabbr{}} & \textbf{No Alignment} & \textbf{Handover Only} & \textbf{\algabbr{}} & \textbf{No Alignment} & \textbf{Handover Only} & \textbf{\algabbr{}} & \textbf{No Alignment} & \textbf{Handover Only} & \textbf{\algabbr{}}  
  \\
\midrule
\textbf{PSNR} $\uparrow$ & 26.52 & 27.36 & \textbf{31.12} & 23.66 & 24.52 & \textbf{26.08} & 25.94 & 24.45 & \textbf{29.26} & 23.02 & 23.95 & \textbf{30.52} & 22.10 & 22.20 & \textbf{28.51}  \\

\textbf{SSIM} $\uparrow$ & 0.991 & 0.991 & \textbf{0.994} & 0.982 & 0.983 & \textbf{0.984} & 0.986 & 0.986 & \textbf{0.989} & 0.985 & 0.984 & \textbf{0.989} & 0.969 & 0.970 & \textbf{0.981}  \\

\textbf{LPIPS} $\downarrow$ & 0.015 & 0.015 & \textbf{0.010} & 0.037 & 0.032 & \textbf{0.025} & 0.019 & 0.018 & \textbf{0.011} & 0.020 & 0.020 & \textbf{0.011} & 0.046 & 0.038 & \textbf{0.020} \\
\bottomrule
\end{tabular}
}
\caption{\textbf{Omnidirectional Object Reconstruction Quality} Comparison of reconstruction quality of 5 home, industrial, and office objects. We report metrics for each object's $\text{3DGS}_\text{merge}$ averaged over \textbf{200} images from the left gripper and right gripper scans. Peak Signal-to-Noise Ratio (PSNR) quantifies the quality of a reconstructed or compressed image/video by comparing it to the original on the logarithmic decibel scale, where higher values indicate better fidelity. Structural Similarity Index (SSIM) measures the similarity between two images by considering luminance, contrast, and structure, with values ranging from -1 to 1, where 1 indicates identical images. Learned Perceptual Image Patch Similarity (LPIPS) measures the perceptual similarity between images by comparing feature embeddings from a pre-trained neural network, ranging from 0 to 1, where lower values indicate higher similarity. Results suggest that \algabbr{} is able to reconstruct objects with high quality by incorporating information from all view directions.}
\label{table:categorized-objects}
\vspace{-2em}
\end{table*}
\begin{table}[!ht]
\label{tab:defects}
\centering
\begin{tabular}{ccc}
\toprule
\textbf{Geometric Defects} & \textbf{Visual Defects} & \textbf{Success Rate} \\
\midrule
6/7 & 5/5 & 83.3\% \\
\bottomrule
\end{tabular}
\caption{\textbf{Correct Identifications} of the defective object using aligned pairwise comparisons.}
\vspace{-2em}
\end{table}
\section{Experiments}

Physical experiments aim to evaluate 1) the quality of the 3D reconstruction, and 2) the effectiveness of the inspection system for finding defects.

\subsection{Reconstruction}

As shown in Fig. \ref{table:categorized-objects}, we collect 17 objects for reconstruction, which comprise a range of industrial, office, and household objects. We evaluate the reconstruction quality by comparing object renderings to the 200 ground truth camera images, reporting image similarity metrics (PSNR, SSIM, and LPIPS) on image regions masked by the intersection of the object mask and the accumulation (excluding the gripper). This penalizes accumulation and shape disparities. $\text{3DGS}_\text{merge}$ is compared to images from the left \textit{and} right hand scans, ensuring that it holistically represents the object. On average, the full Omni-Scan pipeline takes approximately 84 minutes per object on a single NVIDIA RTX 4090 GPU: 9 minutes for scanning (including left scan, handover, and right scan), 46 minutes for object mask generation (200 images total from both grippers), and 28 minutes for 3DGS training (including individual left/right models, and final merged model).

\paragraph{Results} See Figure~\ref{fig:reconstructions} for qualitative multi-view renders of objects reconstructed autonomously by \algabbr{}. Table~\ref{table:categorized-objects} reports image quality metrics across both left and right datasets. \algabbr{} achieves high reconstruction quality, indicating it is able to reconstruct even occluded regions of the object by incorporating information from the un-occluded dataset. 

\begin{figure}
\vspace{-0.5em}
    \centering
    \includegraphics[width=0.9\linewidth]{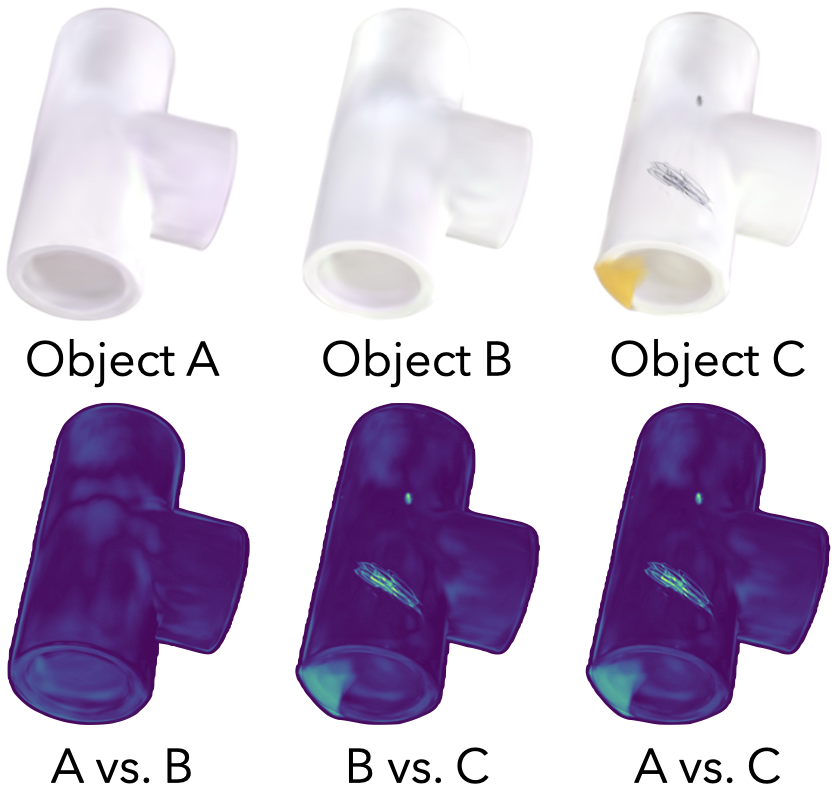}
    \caption{\textbf{Visual Defect Detection} \textit{Top Row} The rendered RGB of three \algabbr models. \textit{Bottom Row}  The colorized per-pixel difference after alignment. The highest difference appears in the exact position of the scratch and tape.} 
    \label{fig:vis_defect}
    \vspace{-2em}
\end{figure}
\begin{figure}
    \centering
    \includegraphics[width=0.9\linewidth]{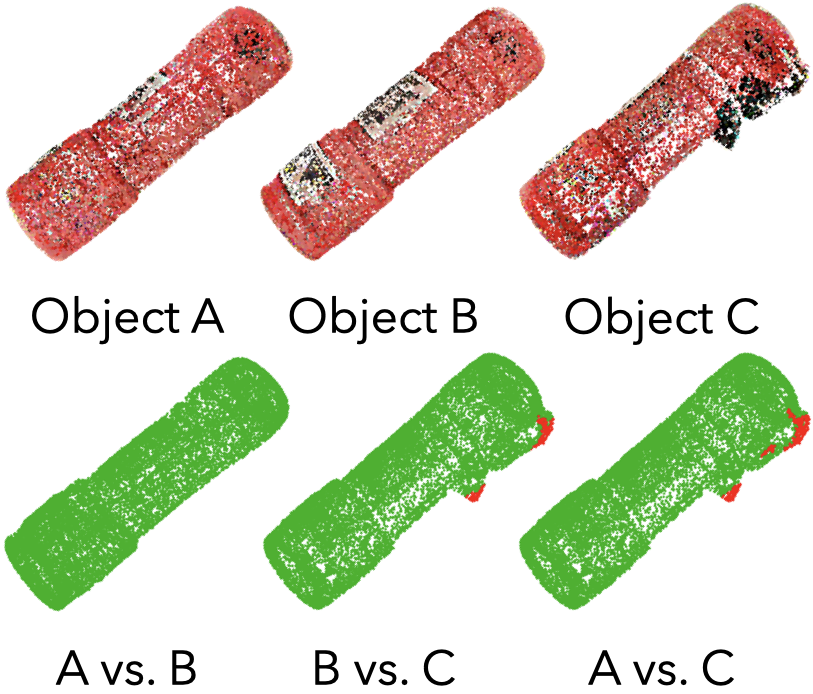}
    \caption{\textbf{Geometric Defect Detection} \textit{Top Row} The aligned point clouds of three scanned objects. \textit{Bottom Row} The point cloud difference between any two point clouds. Green points are points that are within the minimum distance to any other point on the other point cloud while red points are points which exceed this threshold and are classified as defect points.}
    \label{fig:geo_defect}
    \vspace{-1.5em}
\end{figure}
\subsection{Defect Inspection}
We apply \algabbr{} for defect inspection on 12 distinct objects, with 3 scans for each object where 2 are of pristine reference objects and 1 contains a visual or geometric defect. 

Visual defects are changes made to the visual appearance of the object without significantly affecting its geometry. For the PVC pipe connector in Figure \ref{fig:vis_defect}, we add yellow tape and mark one end of the pipe. 
Geometric defects are introduced by damaging or otherwise changing the surface geometry of the object. For example, in Figure \ref{fig:geo_defect} we attach a strap to the end of the flashlight but changes such as bending, breaking, or cutting the object also qualify. 

We evaluate the system's ability to identify the defective part of these 3 scans. \algabbr{} highlights the point clouds of physical defects and highlights renders of a difference visual defects.
We identify the defective part using a combination of pixel-space analysis and point cloud analysis. 
We use TEASER++ \cite{Yang20tro-teaser},  \cite{5152473}, a fast and robust global registration method, to obtain an initial alignment transformation between the extracted point clouds. This transformation serves as an initialization for ICP, which further refines the alignment between the Gaussian models. 

\paragraph{Pixel Differencing} 
We render 100 images from poses that align with the training dataset for the first dataset. Then using the alignment transform of the following 2 datasets, we compute renders of the same location and orientation. We can then compute pixel the difference of these two renders to evaluate the difference of the models. Since the two non-defective objects should be indistinguishable, we can compare pair-wise L2 distances between the RGB values. Specifically, the average distance across all the rendered frames of the two scans for each frame is computed. The smallest distance pair is the non-defective parts with the remainder being the defective one as demonstrated in Fig. \ref{fig:vis_defect}.

\paragraph{Pixel Differencing Results}
\algabbr successfully detects visual defects in 4 out of 5 trials. We successfully identified defects such as scratches and tape on the pipe connector as illustrated in Fig. \ref{fig:vis_defect}. Results suggest that our alignment pipeline correctly aligns the different scans and direct comparison of the aligned pixels is a valid way to perform defect detection. 


\paragraph{Point Cloud Differencing} Given the aligned point clouds for any two objects, we compute the difference between them. This is done by computing the minimum distance from a point in one point cloud to any point in the other point cloud. If a point's minimum distance from any other point exceeds our distance threshold of 4.5mm (empirically determined based on our set of objects to cause no false positives), then we classify it as a defective point.  

\paragraph{Geometric Defect Detection Results} \algabbr is able to correctly identify the geometric defect in 6 out of 7 trials. These results indicate that the point clouds generated by training a \algabbr are quite consistent among different undamaged objects as they have next to no defect points that exceeded our distance threshold of 4.5mm. This also further reinforces the ability of the alignment pipeline to properly align these models. Point cloud differencing fails on the pressure sensor with a geometric defect of slight sanding on one end of the object and a cut made on another end. These defects are marginal and the resulting point cloud does not noticeably differ from the two reference object point clouds.  



\section{Limitations}
One limitation of \algabbr{} is with specularities. When scanning metallic objects, the color as well as the brightness can change depending on the pose of the camera to the object. This leads to issues with alignment and pixel differencing, since the same point on the object may look very different to the model depending on how it was grasped/ scanned.
The system also relies on the handover pose as a good initialization for the Iterative Closest Point to estimate the transform between the left and right datasets. If the object slips significantly during handover, the resulting pose estimation ceases to be accurate, and the overall model quality suffers as a result. Since 3DGS models can contain gaussians in their interior, geometric differencing sometimes presents spurious false positives. Future work will explore mesh-based approaches for geometric differencing that better localize geometric defects.
Additionally, the masking processing time may be accelerated by parallelizing the processing across multiple servers or GPUs, or by recent work \cite{pfaff2025scalablereal2simphysicsawareasset}, \cite{ren2024grounding}, and \cite{ren2024grounded}. The empirically determined thresholds for masking could be replaced by thresholds optimized through learning-based methods. Similarly, using learning-based similarity models, such as \cite{chen2020simple}, could improve defect detection.
\section{Conclusion}
In this paper, we present \algabbr{}, a system for autonomous high-quality robotic creation of omni-directional digital twins and defect inspection. Experiments suggest that \algabbr{} constructs models with sufficient visual fidelity to detect visual and geometric defects on household, office, and industrial objects with up to 83.3\% accuracy.

\bibliographystyle{IEEEtran}
\bibliography{references}

\begin{thebibliography}{10}
\providecommand{\url}[1]{#1}
\csname url@samestyle\endcsname
\providecommand{\newblock}{\relax}
\providecommand{\bibinfo}[2]{#2}
\providecommand{\BIBentrySTDinterwordspacing}{\spaceskip=0pt\relax}
\providecommand{\BIBentryALTinterwordstretchfactor}{4}
\providecommand{\BIBentryALTinterwordspacing}{\spaceskip=\fontdimen2\font plus
\BIBentryALTinterwordstretchfactor\fontdimen3\font minus \fontdimen4\font\relax}
\providecommand{\BIBforeignlanguage}[2]{{%
\expandafter\ifx\csname l@#1\endcsname\relax
\typeout{** WARNING: IEEEtran.bst: No hyphenation pattern has been}%
\typeout{** loaded for the language `#1'. Using the pattern for}%
\typeout{** the default language instead.}%
\else
\language=\csname l@#1\endcsname
\fi
#2}}
\providecommand{\BIBdecl}{\relax}
\BIBdecl

\bibitem{mildenhall2020nerf}
B.~Mildenhall, P.~P. Srinivasan, M.~Tancik, J.~T. Barron, R.~Ramamoorthi, and R.~Ng, ``Nerf: Representing scenes as neural radiance fields for view synthesis,'' in \emph{ECCV}, 2020.

\bibitem{kerbl2023gaussian}
B.~Kerbl, G.~Kopanas, T.~Leimk{\"u}hler, and G.~Drettakis, ``3{D} {G}aussian splatting for real-time radiance field rendering,'' \emph{ACM Transactions on Graphics}, 2023.

\bibitem{NEURIPS2024_26cfdcd8}
L.~Yang, B.~Kang, Z.~Huang, Z.~Zhao, X.~Xu, J.~Feng, and H.~Zhao, ``Depth anything v2,'' in \emph{Advances in Neural Information Processing Systems}, vol.~37, 2024.

\bibitem{kirillov2023segment}
A.~Kirillov, E.~Mintun, N.~Ravi, H.~Mao, C.~Rolland, L.~Gustafson, T.~Xiao, S.~Whitehead, A.~C. Berg, W.-Y. Lo \emph{et~al.}, ``Segment anything,'' in \emph{ICCV}, 2023.

\bibitem{ravi2025sam}
N.~Ravi, V.~Gabeur, Y.-T. Hu, R.~Hu, C.~Ryali, T.~Ma, H.~Khedr, R.~R{\"a}dle, C.~Rolland, L.~Gustafson, E.~Mintun, J.~Pan, K.~V. Alwala, N.~Carion, C.-Y. Wu, R.~Girshick, P.~Dollar, and C.~Feichtenhofer, ``{SAM} 2: Segment anything in images and videos,'' in \emph{The Thirteenth International Conference on Learning Representations}, 2025.

\bibitem{adamkiewicz2022vision}
M.~Adamkiewicz, T.~Chen, A.~Caccavale, R.~Gardner, P.~Culbertson, J.~Bohg, and M.~Schwager, ``Vision-only robot navigation in a neural radiance world,'' \emph{IEEE Robotics and Automation Letters}, 2022.

\bibitem{barron2021mip}
J.~T. Barron, B.~Mildenhall, M.~Tancik, P.~Hedman, R.~Martin-Brualla, and P.~P. Srinivasan, ``Mip-nerf: A multiscale representation for anti-aliasing neural radiance fields,'' in \emph{ICCV}, 2021.

\bibitem{barron2022mip}
J.~T. Barron, B.~Mildenhall, D.~Verbin, P.~P. Srinivasan, and P.~Hedman, ``Mip-nerf 360: Unbounded anti-aliased neural radiance fields,'' in \emph{CVPR}, 2022.

\bibitem{ma2022deblur}
L.~Ma, X.~Li, J.~Liao, Q.~Zhang, X.~Wang, J.~Wang, and P.~V. Sander, ``Deblur-nerf: Neural radiance fields from blurry images,'' in \emph{CVPR}, 2022.

\bibitem{tancik2023nerfstudio}
M.~Tancik, E.~Weber, E.~Ng, R.~Li, B.~Yi, T.~Wang, A.~Kristoffersen, J.~Austin, K.~Salahi, A.~Ahuja \emph{et~al.}, ``Nerfstudio: A modular framework for neural radiance field development,'' in \emph{ACM SIGGRAPH 2023 conference proceedings}, 2023.

\bibitem{wang2023f2}
P.~Wang, Y.~Liu, Z.~Chen, L.~Liu, Z.~Liu, T.~Komura, C.~Theobalt, and W.~Wang, ``F2-nerf: Fast neural radiance field training with free camera trajectories,'' in \emph{CVPR}, 2023.

\bibitem{barron2023zip}
J.~T. Barron, B.~Mildenhall, D.~Verbin, P.~P. Srinivasan, and P.~Hedman, ``Zip-nerf: Anti-aliased grid-based neural radiance fields,'' in \emph{ICCV}, 2023, pp. 19\,697--19\,705.

\bibitem{muller2022instant}
T.~M{\"u}ller, A.~Evans, C.~Schied, and A.~Keller, ``Instant neural graphics primitives with a multiresolution hash encoding,'' \emph{ACM Transactions on Graphics (ToG)}, vol.~41, no.~4, pp. 1--15, 2022.

\bibitem{Chen2022ECCV}
A.~Chen, Z.~Xu, A.~Geiger, J.~Yu, and H.~Su, ``Tensorf: Tensorial radiance fields,'' in \emph{ECCV}, 2022.

\bibitem{fridovich2023k}
S.~Fridovich-Keil, G.~Meanti, F.~R. Warburg, B.~Recht, and A.~Kanazawa, ``K-planes: Explicit radiance fields in space, time, and appearance,'' in \emph{CVPR}, 2023, pp. 12\,479--12\,488.

\bibitem{fridovich2022plenoxels}
S.~Fridovich-Keil, A.~Yu, M.~Tancik, Q.~Chen, B.~Recht, and A.~Kanazawa, ``Plenoxels: Radiance fields without neural networks,'' in \emph{CVPR}, 2022.

\bibitem{park2021hypernerf}
K.~Park, U.~Sinha, P.~Hedman, J.~T. Barron, S.~Bouaziz, D.~B. Goldman, R.~Martin-Brualla, and S.~M. Seitz, ``Hypernerf: A higher-dimensional representation for topologically varying neural radiance fields,'' \emph{ACM Trans. Graph.}, vol.~40, no.~6, dec 2021.

\bibitem{li2023dynibar}
Z.~Li, Q.~Wang, F.~Cole, R.~Tucker, and N.~Snavely, ``Dynibar: Neural dynamic image-based rendering,'' in \emph{CVPR}, 2023.

\bibitem{pumarola2020d}
A.~Pumarola, E.~Corona, G.~Pons-Moll, and F.~Moreno-Noguer, ``{D-NeRF: Neural Radiance Fields for Dynamic Scenes},'' in \emph{CVPR}, 2020.

\bibitem{Zhu_2022_CVPR}
Z.~Zhu, S.~Peng, V.~Larsson, W.~Xu, H.~Bao, Z.~Cui, M.~R. Oswald, and M.~Pollefeys, ``Nice-slam: Neural implicit scalable encoding for slam,'' in \emph{CVPR}, June 2022.

\bibitem{Sucar_2021_ICCV}
E.~Sucar, S.~Liu, J.~Ortiz, and A.~J. Davison, ``imap: Implicit mapping and positioning in real-time,'' in \emph{ICCV}, October 2021.

\bibitem{rosinol2023nerf}
A.~Rosinol, J.~J. Leonard, and L.~Carlone, ``Nerf-slam: Real-time dense monocular slam with neural radiance fields,'' in \emph{IROS}.\hskip 1em plus 0.5em minus 0.4em\relax IEEE, 2023.

\bibitem{li20223d}
Y.~Li, S.~Li, V.~Sitzmann, P.~Agrawal, and A.~Torralba, ``3d neural scene representations for visuomotor control,'' in \emph{CoRL}.\hskip 1em plus 0.5em minus 0.4em\relax PMLR, 2022.

\bibitem{22-driess-NeRF-RL}
D.~Driess, I.~Schubert, P.~Florence, Y.~Li, and M.~Toussaint, ``Reinforcement learning with neural radiance fields,'' in \emph{NeurIPS}, 2022.

\bibitem{kerr2022evonerf}
J.~Kerr, L.~Fu, H.~Huang, Y.~Avigal, M.~Tancik, J.~Ichnowski, A.~Kanazawa, and K.~Goldberg, ``Evo-ne{RF}: Evolving ne{RF} for sequential robot grasping of transparent objects,'' in \emph{CoRL}, 2022.

\bibitem{IchnowskiAvigal2021DexNeRF}
J.~Ichnowski*, Y.~Avigal*, J.~Kerr, and K.~Goldberg, ``{Dex-NeRF}: Using a neural radiance field to grasp transparent objects,'' in \emph{CoRL}, 2020.

\bibitem{Rashid2023LanguageER}
A.~Rashid, S.~Sharma, C.~M. Kim, J.~Kerr, L.~Y. Chen, A.~Kanazawa, and K.~Goldberg, ``Language embedded radiance fields for zero-shot task-oriented grasping,'' in \emph{CoRL}, 2023.

\bibitem{kerr2024rsrd}
J.~Kerr, C.~M. Kim, M.~Wu, B.~Yi, Q.~Wang, K.~Goldberg, and A.~Kanazawa, ``Robot see robot do: Imitating articulated object manipulation with monocular 4d reconstruction,'' in \emph{CoRL}, 2024.

\bibitem{shen2023F3RM}
W.~Shen, G.~Yang, A.~Yu, J.~Wong, L.~P. Kaelbling, and P.~Isola, ``Distilled feature fields enable few-shot language-guided manipulation,'' in \emph{CoRL}, 2023.

\bibitem{byravan2023nerf2real}
A.~Byravan, J.~Humplik, L.~Hasenclever, A.~Brussee, F.~Nori, T.~Haarnoja, B.~Moran, S.~Bohez, F.~Sadeghi, B.~Vujatovic \emph{et~al.}, ``Nerf2real: Sim2real transfer of vision-guided bipedal motion skills using neural radiance fields,'' in \emph{ICRA}.\hskip 1em plus 0.5em minus 0.4em\relax IEEE, 2023.

\bibitem{downs2022google}
L.~Downs, A.~Francis, N.~Koenig, B.~Kinman, R.~Hickman, K.~Reymann, T.~B. McHugh, and V.~Vanhoucke, ``Google scanned objects: A high-quality dataset of 3d scanned household items,'' in \emph{ICCV}, 2022.

\bibitem{aanaes2016large}
H.~Aan{\ae}s, R.~R. Jensen, G.~Vogiatzis, E.~Tola, and A.~B. Dahl, ``Large-scale data for multiple-view stereopsis,'' \emph{International Journal of Computer Vision}, pp. 1--16, 2016.

\bibitem{deitke2023objaverse}
M.~Deitke, D.~Schwenk, J.~Salvador, L.~Weihs, O.~Michel, E.~VanderBilt, L.~Schmidt, K.~Ehsani, A.~Kembhavi, and A.~Farhadi, ``Objaverse: A universe of annotated 3d objects,'' in \emph{CVPR}, 2023, pp. 13\,142--13\,153.

\bibitem{zhong2024color}
L.~Zhong, L.~Yang, K.~Li, H.~Zhen, M.~Han, and C.~Lu, ``Color-neus: Reconstructing neural implicit surfaces with color,'' in \emph{2024 International Conference on 3D Vision (3DV)}.\hskip 1em plus 0.5em minus 0.4em\relax IEEE, 2024, pp. 631--640.

\bibitem{wen2023bundlesdf}
B.~Wen, J.~Tremblay, V.~Blukis, S.~Tyree, T.~M{\"u}ller, A.~Evans, D.~Fox, J.~Kautz, and S.~Birchfield, ``{BundleSDF}: Neural 6-dof tracking and 3d reconstruction of unknown objects,'' in \emph{CVPR}.\hskip 1em plus 0.5em minus 0.4em\relax IEEE, 2023.

\bibitem{khan2021vision}
A.~Khan, C.~Mineo, G.~Dobie, C.~Macleod, and G.~Pierce, ``Vision guided robotic inspection for parts in manufacturing and remanufacturing industry,'' \emph{Journal of Remanufacturing}, vol.~11, no.~1, pp. 49--70, 2021.

\bibitem{davtalab2022automated}
O.~Davtalab, A.~Kazemian, X.~Yuan, and B.~Khoshnevis, ``Automated inspection in robotic additive manufacturing using deep learning for layer deformation detection,'' \emph{Journal of Intelligent Manufacturing}, vol.~33, no.~3, pp. 771--784, 2022.

\bibitem{elgeneidy2019gripper}
K.~Elgeneidy, P.~Lightbody, S.~Pearson, and G.~Neumann, ``Characterising 3d-printed soft fin ray robotic fingers with layer jamming capability for delicate grasping,'' in \emph{2019 2nd IEEE International Conference on Soft Robotics (RoboSoft)}, 2019, pp. 143--148.

\bibitem{shankar2022learned}
K.~Shankar, M.~Tjersland, J.~Ma, K.~Stone, and M.~Bajracharya, ``A learned stereo depth system for robotic manipulation in homes,'' \emph{IEEE Robotics and Automation Letters}, vol.~7, no.~2, 2022.

\bibitem{ester1996density}
M.~Ester, H.-P. Kriegel, J.~Sander, X.~Xu \emph{et~al.}, ``A density-based algorithm for discovering clusters in large spatial databases with noise.'' in \emph{kdd}, vol.~96, no.~34, 1996, pp. 226--231.

\bibitem{sundermeyer2021contact}
M.~Sundermeyer, A.~Mousavian, R.~Triebel, and D.~Fox, ``Contact-graspnet: Efficient 6-dof grasp generation in cluttered scenes,'' in \emph{ICRA}, 2021.

\bibitem{jacobi2024motion}
I.~Jacobi~Robotics, ``Jacobi motion library -- next generation motion planning,'' 2024, https://docs.jacobirobotics.com.

\bibitem{nerfstudio}
M.~Tancik, E.~Weber, E.~Ng, R.~Li, B.~Yi, J.~Kerr, T.~Wang, A.~Kristoffersen, J.~Austin, K.~Salahi, A.~Ahuja, D.~McAllister, and A.~Kanazawa, ``Nerfstudio: A modular framework for neural radiance field development,'' in \emph{ACM SIGGRAPH 2023 Conference Proceedings}, ser. SIGGRAPH '23, 2023.

\bibitem{ye2024gsplatopensourcelibrarygaussian}
V.~Ye, R.~Li, J.~Kerr, M.~Turkulainen, B.~Yi, Z.~Pan, O.~Seiskari, J.~Ye, J.~Hu, M.~Tancik, and A.~Kanazawa, ``gsplat: An open-source library for {Gaussian} splatting,'' \emph{Journal of Machine Learning Research}, 2024.

\bibitem{Yang20tro-teaser}
H.~Yang, J.~Shi, and L.~Carlone, ``{TEASER: Fast and Certifiable Point Cloud Registration},'' \emph{{IEEE} Trans. Robotics}, 2020.

\bibitem{5152473}
R.~B. Rusu, N.~Blodow, and M.~Beetz, ``Fast point feature histograms (fpfh) for 3d registration,'' in \emph{ICRA}, 2009, pp. 3212--3217.

\bibitem{pfaff2025scalablereal2simphysicsawareasset}
N.~Pfaff, E.~Fu, J.~Binagia, P.~Isola, and R.~Tedrake, ``Scalable real2sim: Physics-aware asset generation via robotic pick-and-place setups,'' \emph{arXiv preprint arXiv:2503.00370}, 2025.

\bibitem{ren2024grounding}
T.~Ren, Q.~Jiang, S.~Liu, Z.~Zeng, W.~Liu, H.~Gao, H.~Huang, Z.~Ma, X.~Jiang, Y.~Chen, Y.~Xiong, H.~Zhang, F.~Li, P.~Tang, K.~Yu, and L.~Zhang, ``Grounding dino 1.5: Advance the "edge" of open-set object detection,'' \emph{arXiv preprint}, 2024.

\bibitem{ren2024grounded}
T.~Ren, S.~Liu, A.~Zeng, J.~Lin, K.~Li, H.~Cao, J.~Chen, X.~Huang, Y.~Chen, F.~Yan, Z.~Zeng, H.~Zhang, F.~Li, J.~Yang, H.~Li, Q.~Jiang, and L.~Zhang, ``Grounded sam: Assembling open-world models for diverse visual tasks,'' \emph{arXiv preprint}, 2024.

\bibitem{chen2020simple}
T.~Chen, S.~Kornblith, M.~Norouzi, and G.~Hinton, ``A simple framework for contrastive learning of visual representations,'' \emph{arXiv preprint arXiv:2002.05709}, 2020.

\end{thebibliography}

\end{document}